# Driving Digital Engineering Integration and Interoperability Through Semantic Integration of Models with Ontologies

Daniel Dunbar[1] | Thomas Hagedorn[1] | Mark Blackburn[1] | John Dzielski[1] | Steven Hespelt[1] | Benjamin Kruse[2] | Dinesh Verma[1] | Zhongyuan Yu[1]

[1]Systems Engineering Research Center (SERC), Stevens Institute of Technology, Hoboken, NJ 07030, USA
[2]Virginia Polytechnic Institute and State University, Blacksburg, VA 24061, USA

**Abstract**

Engineered solutions are becoming more complex and multi-disciplinary in nature. This evolution requires new techniques to enhance design and analysis tasks that incorporate data integration and interoperability across various engineering tool suites spanning multiple domains at different abstraction levels. Semantic Web Technologies (SWT) offer data integration and interoperability benefits as well as other opportunities to enhance reasoning across knowledge represented in multiple disparate models. This paper introduces the Digital Engineering Framework for Integration and Interoperability (DEFII) for incorporating SWT into engineering design and analysis tasks. The framework includes three notional interfaces for interacting with ontology-aligned data. It also introduces a novel Model Interface Specification Diagram (MISD) that provides a tool-agnostic model representation enabled by SWT that exposes data stored for use by external users through standards-based interfaces. Use of the framework results in a tool-agnostic authoritative source of truth spanning the entire project, system, or mission.

**Significance and Practitioner Points**

This paper develops the use of ontologies and semantic web technologies for Digital Engineering by introducing the Digital Engineering Framework for Integration and Interoperability (DEFII). The DEFII framework establishes three notional interfaces for populating, interacting with, and enhancing ontology-aligned data. It provides the ability to establish tool-agnostic interfaces for data contained in a system design with the novel Model Interface Specification Diagram (MISD). The MISD uses the SysML language as a descriptive model of the analysis system and co-mingles this with the system model, giving modelers the ability to specify interfaces used for analysis in a context dependent manner. This allows for models to be defined to include data from disparate sources within the system and exposed in a way that gives external tools, from industry established design and analysis software to in-house visualization tools, access to an established, authoritative source of truth. This approach to interoperability allows practitioners to continue using tools that suit their needs and preferences while taking advantage of enhanced interoperability between disciplines and creating a knowledge base that can be expanded on in the future to allow for further integration of artificial and augmented intelligence applications.

## 1 | INTRODUCTION

Increasing complexity in engineered projects requires a high level of collaboration across disciplines. To maintain high standards of quality and reasonable time frames, computer assisted collaboration is increasingly necessary. Data integration that enables cross-domain reasoning and collaboration at the model level is key to enabling and enhancing computer assisted engineering and design across multiple abstraction levels, domains, and disciplines.

Digital Engineering (DE) is an umbrella domain that seeks to integrate the Model Based Systems Engineering (MBSE) and domain specific Model Based Engineering (MBE) domains. Defense Acquisition University defines DE as "an integrated digital approach that uses authoritative sources of systems' data and models as a continuum across disciplines to support lifecycle activities from concept through disposal."[1] In order to do this, it relies heavily on the concepts of Authoritative Source of Truth (AST) and Digital Thread (DT). The AST serves as a central data registry where all tools associated with the project must go to access system data. Thus, multiple tools can access the same datapoint and have certainty that they are working from the same value. Researchers have used various structures to form the AST, including system models[2], databases[3], and various graph data structures[4–7]. The DT digitally connects different models used in system design and analysis across domain and software boundaries.[8] This requires a robust data integration plan to enable the model level connection that is needed.

Integration of various models is traditionally done through tool-to-tool integration. Custom interfaces designed to connect tool A to tool B have been used for years to accommodate data integration needs in cross domain analysis. However, tool-to-tool integration is brittle in nature[4], and the number of tool-to-tool integrations needed to fully connect the design and analysis system grows exponentially as tools are added[9]. In other words, tool-to-tool integration is prone to error, difficult to maintain, and hard to scale.

A database approach, which allows for a central hub of data that is accessed by multiple tools, is a better fit as a DE AST. Particularly, a graph data structure that captures system specific data and connects it to domain knowledge presents an interesting foundation for a potential solution. Graph data structures capture relationships between nodes and can take advantage of both relational and graph based algorithms.[10]

Computer based ontologies and associated technologies collectively known as Semantic Web Technologies (SWT) provide a technological base to build out a graph data structure as an AST in the DE context. SWT comprises a suite of technologies to tag graph data with semantically meaningful labels, to store and retrieve that data, and to automatically reason upon it based upon the logical framework underpinning the tags. Ontologies provide the markup schema used to tag the graph and a logical formalization in the form of a taxonomy of terms, relations, and machine-readable logical expressions using them. Data repositories called triple stores provide storage and endpoints for semantic queries, which are used to retrieve and interact with graph data. Automated reasoners afford the ability to deduce new relations within tagged data based upon logical inferencing, and to enhance query results with that inferred information.

While graph data structures have been used in DE and systems engineering domains with positive results[4–7], to date there is not a robust framework for guiding engineers to integrate SWT into a DE environment.

This paper presents the Digital Engineering Framework for Integration and Interoperability (DEFII), an ontology-based framework for solving integration and interoperability concerns inherent to the DE domain space. It uses a graph data structure as the DE AST and establishes three notional interfaces to allow interaction with the ontology aligned data: the Direct Interface, the Mapping Interface, and the Specified Model Interface. Additionally, the paper introduces a novel Model Interface Specification Diagram (MISD) to specify ontologically relevant interfaces without deep understanding of the ontologies being used.

A successful framework for integrating SWT into DE must move beyond theory and enable practice that realizes these benefits. For a framework to reach that threshold, this paper holds that it must:
1. Provide clear avenues for mapping data from engineering design and analysis models to an ontology-aligned data store
2. Allow access to contained data in a flexible, tool-agnostic manner
3. Allow for the transformation and enhancement of data using various semantic technologies

These three success criteria address current data integration needs in DE and prepare for future capability by taking advantage of an ontology-based AST that provides a holistic view of the systems under design. This paper shows the DEFII framework satisfies all three criteria.

Section 2 looks at related work and identifies a gap the DEFII framework fills in existing literature. Section 3 details the DEFII framework and notional interfaces related to it. It also introduces a case study from the Information Technology (IT) cyber security domain that uses DEFII to produce results that can be analyzed. Section 4 instantiates the notional interfaces to address a specific use case and describes the results of usage of the DEFII framework. Section 5 analyzes the results to validate DEFII against the success criteria defined above, discusses limitations of the research and opportunities to extend the research, and interprets the results for the larger DE context. Section 6 provides a conclusion to the paper.

## 2 | RELATED WORK

Ontologies have long been proposed as a medium for knowledge representation in engineering. Theoretical benefits include the potential for reuse, automated inferencing, and knowledge sharing[11–13]. . These theoretical benefits have led to research on how to build solutions that make use of the SWT to enhance engineering efforts. A first step in this research is the use of SWT directly.

A significant body of past research has focused on demonstrating engineering capabilities that are enabled by SWT. Most prior work focuses narrowly on either pure ontology definition for the engineering domain or the capabilities afforded by ontology aligned data. Coelho et al. use direct invocation of the SWT stack in their proposal of the "Data-Ontology-Rule footing" as a mechanism for building ontology integration into the design and analysis workflow. Mechanisms for pulling system data into ontologies beyond direct manipulation of the ontological data using semantic tools are not discussed[14]. Eddy et al. describe a framework for design alternative development based on the use of modular ontologies of the engineering design domain and various reasoning capabilities[15]. Similarly, Hagedorn et al. use SWT to enable design and ideation in the Additive Manufacturing domain[16] based upon semantic querying and rule-based inferences. Daun et al. use ontologies to describe contexts in concurrent engineering[17]. While these works demonstrate possible benefits of SWT in engineering context, they lack defined techniques for populating ontology aligned-data instances and interacting with engineering models or tools. Thus the benefits demonstrated in these and similar works remain largely hypothetical [6].

Several works have attempted to bridge this gap between tools and SWT using a process of data mapping. Termed "data ingestion" by commercial tools[18], mapping is a process by which data from outside the SWT stack is parsed into SWT formats and aligned to ontology-tagged graph patterns. El Kadiri and Kiritsis discuss mapping for ontology use in the Product Lifecycle Management (PLM) domain[19]. Several other publications discuss mapping both specifically and generically in reference to systems engineering domain[10,18,20–23]. Mapping tool and model information to ontology-aligned data allows the use of the SWT in more realistic engineering workflows using real or realistic data. Examples of this include JPL's use in CEASAR to establish logical consistency and build reports through the use of custom queries[24], Lu et al.'s use of custom querying[21], DL reasoning by Petnga et al.[23], and data exploration by Hagedorn et al.[18].

Multiple frameworks have been introduced that make use of direct mappings from tools to enable SWT applications in engineering contexts. NASA's Jet Propulsion Lab (JPL) has introduced the Computer Aided Engineering for Spacecraft Architectures Tool Suite (CAESAR) as a type of framework to aid in MBSE. CAESAR uses the Ontological Modeling Language (OML) to capture models in a controlled vocabulary and a number of tool adapters to incorporate different external design and analysis tools into the ontology-based approach[24]. Moser has proposed the Engineering Knowledge Base (EKB) as a framework for integrating SWT in multidisciplinary engineering environments[9]. This framework focuses on the use of custom tool mappings to interface with the underlying graph data structure[9,25]. Moser contrasts this approach with a common repository approach. In a common repository approach, schema must be predetermined and are tool specific, which makes them more fragile and harder to maintain. In the EKB approach, tool data is mapped to common engineering concepts, which enables a more robust representation of the data and semi-automatic transformation of data from one tool to another[9]. Both the CAESAR and EKB frameworks use an ontology-based AST and enable aspects of the DT. Mapping is often used alongside descriptive models such as those defined using the Systems Modeling Language (SysML). NASA's Jet Propulsion Lab (JPL)[5,26], Bone et al.[7], and Blackburn et al.[27] all populate SWT tools with ontologically aligned information using SysML that use stereotypes to inject ontological tags. In all these cases, stereotypes serve to guide a tool-specific mapping process.

Across all of these efforts, mapping provides a mechanism for connecting engineering tools with ontology-aligned data and the broader SWT stack. However, it does so through a rigid, often tool-specific, connection point. As a result, it may be difficult or labor intensive to add new tools to a workflow. It also limits accessibility the data to those with knowledge of how to use SWT tools. Broader access, through a more flexible input and output mechanism would enable more design and analysis functionality. SysML v2 begins to provide more flexible access to system data through the use of a standards based Application Programming Interface (API)[28,29]. However, SysML v2 uses this flexible input/output mechanism to access the system model, not an ontology-aligned representation of the information stored in the system model.

DTs require the interaction of multiple tools external to an AST. Flexibility in how tools can interact with data in the AST enables more opportunity for this interaction and increases the usability of a DE solution. While the SWT stack and mapping provide capability for a DT and are being used in current engineering research, there is space for additional types of interfaces to enable more diverse access to the AST.

## 3 | METHODS

The DEFII framework structures usage of SWT in the DE context. This section will define the framework and introduce a case study from the Information Technology (IT) domain to validate the usability of the framework in a domain setting.

### 3.1 | Framework Description

The DEFII framework (Figure 1) assigns the role of the AST to ontologies and data aligned to those ontologies. This forms the foundation of the framework. It then uses automated reasoning capabilities of SWT to enrich the ontology-aligned data through the use of rules and relationships defined in the ontologies. Finally, it provides clear categories of interfaces that users can use to access, modify, and populate the AST.

**Figure 1** The DEFII Framework for use of SWT in DE contexts

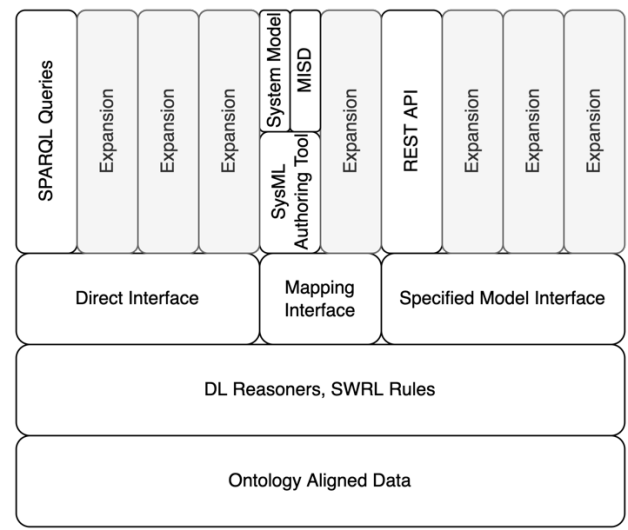

### 3.1.1 | Ontology Aligned Data

Ontology-aligned data is the foundation of the framework. It enables the use of a graph data structure to act as the AST. Using a triple store as a graph repository, the ontological knowledge base uses controlled vocabulary defined by ontology class data to characterize system data as instances of assigned classes. Representation of system data as ontology-aligned data in a graph data structure has four primary benefits:
1. Interoperability

Gruber argues for the use of ontologies for both reusability and interoperability between domains[11,12]. The framework specifies the use of a top-level ontology and shared development principles to drive co-development terminology across disparate knowledge domains. A top-level ontology provides a high-level philosophical basis and development guidance that is shared by all ontologies within an ecosystem aligned to it. Subsequent, domain specific ontologies simply extend this high-level understanding to ever more specific knowledge domains.

2. Tool Agnostic

Because ontologies model domains of knowledge, rather than specific tools or datasets, data aligned to ontologies is tool agnostic. The use of a standards-based open formats, such as the Resource Description Framework (RDF) syntax and the Web Ontology Language (OWL) enhances this tool agnostic representation of system data. This makes the data portable and offers greater flexibility and freedom with tool access to the data.

3. Domain Agnostic

Not only does the graph data structure promote tool agnostic access to data, it also promotes a domain agnostic approach. As noted by Mordecai et al.[10], the use of graph data structures extracts the system representation from the MBSE toolset and presents it in more general terms. As DE serves as an umbrella domain to connect many different domains[30], representation of data in a way that separates it from any specific domain, including the systems engineering domain, offers more equitable access to it.

4. Access to SWT stack

The SWT stack offers many functions based on the formal nature of ontology-aligned data and the triple format in which data is stored. Access to powerful querying, reasoning, and validation is enabled by the foundational decision of the framework to use ontology-aligned data and will be detailed in subsequent sections of the paper.

### 3.1.2 | Reasoning Layer

The reasoning layer uses some of the automated reasoning capabilities included in the SWT stack. Since ontologies make use of axioms and relationships to characterize classes within them, these axioms can be used to further enrich the data beyond what has been explicitly defined. For example, if a child's mother and the mother's mother are defined explicitly in a graph repository and definitions and relations are encoded in the ontology (i.e., a grandmother is a role filled by a parent's mother for that parent's child), then the relationship between the child and the grandmother can be made through automated reasoning, bypassing the need to explicitly declare all knowledge in the graph repository. In characterizing engineering knowledge applied to a system, the ability to infer knowledge based on a heterogenous data store opens the possibility of discovering insights to how design elements originating in separate tools relate to each other. This capability uses mathematical logic, specifically Description Logics (DL), to enable DL reasoners as part of the SWT stack to automatically enrich the data without any extra involvement by external users.

### 3.1.3 | Notional Interfaces

The characterization of the system data in the AST is established, updated, viewed, and analyzed through interfaces to external sources. DEFII specifies three types of notional interfaces: the Direct Interface, the Mapping Interface, and the Specified Model Interface. The Specified Model Interface is further refined by the introduction of the Model Interface Specification Diagram (MISD). With these three types of interfaces and the MISD, the framework provides a structured approach to access and manipulation of ontology-aligned data.

The Direct Interface enables the use of the SWT stack directly on stored data. Data stored as ontology-aligned data enables the use of SWT tools to extract specific knowledge, apply constraint checking, and more using a growing set of tools in the SWT suite. This interface acknowledges this reality and codifies it within the framework's view of interfaces. While implementation of subsequent interface types will inevitably use SWT tools to enable them, this interface type is distinguished by its direct invocation of those tools.

The Mapping Interface transforms model data into ontology-aligned data by accessing data stored in a model or tool and mapping it to ontological classes in the controlled vocabulary. Beginning with external model data and establishing a connection to an ontology limits this interface type to a tool or model specific implementation per instance of the interface connected. Primarily, this interface addresses MBSE models and maps system models to ontology-aligned data. System models are broad in scope and capture design criteria for multiple domains. Thus, tagging data with relevant ontological terms and mapping that data to an ontology-aligned graph provides flexibility for the system model to be built to describe the relevant domains instead of fitting it into predetermined structure. It also means that the interface is domain agnostic: the ontological tags are simply changed or expanded to handle new domains.

This interface is the most restrictive interface as it is responsible for accessing data stored in other tools, making most instances tool-specific implementations. However, this form of interface still provides reductions in development efforts and maintains the benefits of ontologies over the use of direct tool-to-tool data integrations (Figure 2). The use of a mapping to an AST provides a limiting principle on the development efforts related to interfaces. Instead of a growing stable of tool-to-tool integrations, which will multiply with each tool added, an AST limits development to a single new interface per tool added to the workflow[9]. In addition, if the AST uses a triple store and ontology-aligned data, these add functionality to the AST by giving access to the reasoning capacities of the SWT and the other two types of notional interfaces.

**Figure 2** Mapping to an AST reduces interface development work

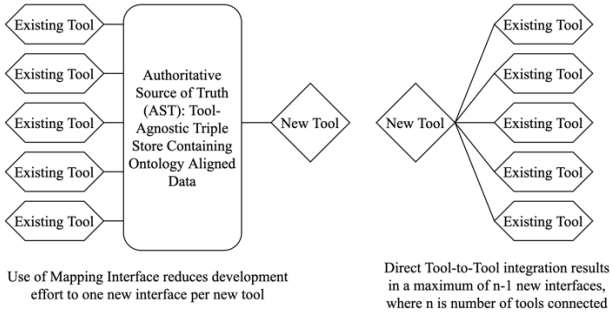

The Specified Model interface characterizes models of interest within the broader system model. A model of interest is defined here as an aggregation of parameters present in the system model that is beneficial for external tools and application purposes. In contrast with the Mapping Interface, the Specified Model Interface begins with ontology-aligned data and exposes this data towards tools. This reverse of direction enables the interface to be tool agnostic. Even if an instantiation of the Specified Model Interface is designed with a particular tool in mind, the direction of the interface creation enables other tools to access the same information via the same interface. This interface primarily addresses MBE models and exposes data in a structured way to be analyzed, visualized, etc.

### 3.1.4 | Model Interface Specification Diagram

The Model Interface Specification Diagram (MISD) is a reusable, graphical specification for the Specified Model Interface. The MISD makes use of SysML parametric diagrams to describe a model of interest and define connections to parameters established elsewhere in the system model (Figure 3). The MISD acts as a graphical specification of data that will be provided to a given tool when requested. This extends Cilli's concept of the Assessment Flow Diagram[31].

Once mapped into the ontological layer, this specification can be concretized in a variety of formats, such as the Comma Separated Value (CSV) format. The MISD allows users unfamiliar with semantic technologies or the underlying ontologies used by the framework to nevertheless request and update ontology-aligned data captured by the system model. It provides the means for tool-agnostic Specified Model Interfaces to be created and accessed by broader user group and toolset.

**Figure 3** Abstract Model Interface Specification Diagram (MISD)

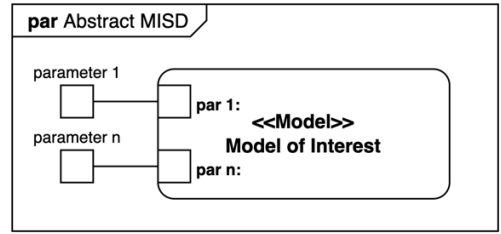

### 3.2 | Framework Case Study

An Information Technology (IT) example is provided as a case study[18,27]. While the DEFII framework is domain agnostic, the IT case study provides a simple use case that is intuitive to understand as a test of the framework and a demonstration of its functionality. The case study examines a simple cyber system that has various cyber elements such as a laptop and software. The system is represented by a SysML Block Definition Diagram (Figure 4).

**Figure 4** BDD Showing Cyber System Architecture Hierarchy

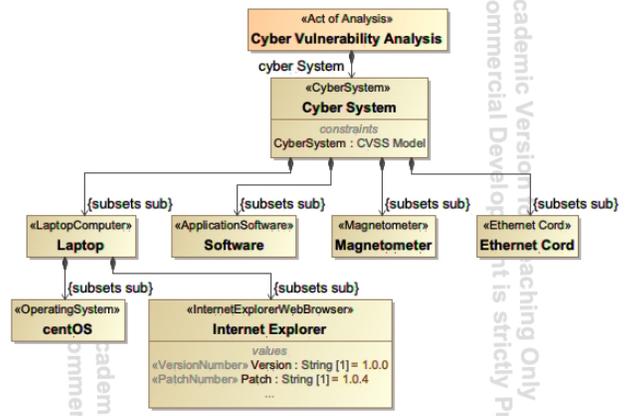

The use case involving this case study is the identification of a seeded cyber vulnerability in the Internet Explorer web browser (contrived for demonstration) and the generation of a system wide Common Vulnerability Scoring System (CVSS)[32] score. This requires the use of both an MBSE model (system model in SysML) and an MBE model (MATLAB analysis model). In addition, a visualization of the results is included to demonstrate that a single tool-agnostic interface can service multiple tools. The DEFII Framework is displayed in the context of the case study (Figure 5). The gray portions of the figure will be realized as results are generated in Section 4.

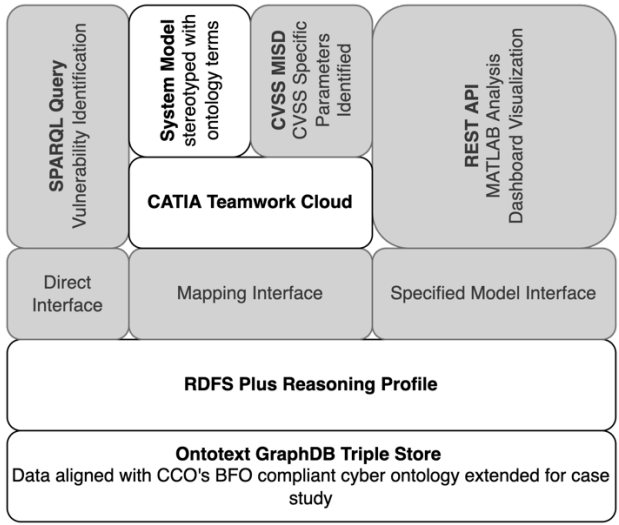

**Figure 5** DEFII Framework applied to cyber case study

To account for the cyber vulnerability portion of the use case, each cyber element also includes the value properties listed in Table I related to a CVSS score. These attributes are used to apply a CVSS score to a specific cyber vulnerability. For example, the scope value can be "Unchanged" or "Changed," and the CVSS scoring process would take that value into account when calculating the overall CVSS score. They are left off Figure 4 for readability.

**Table I** List of CVSS related value properties assigned to each block

| ac (Attack Complexity) | av (Attack Vector) |
|---|---|
| a (Availability) | vs (CVSS Vector) |
| score (CVSS Overall Score) | i (Integrity) |
| s (Scope) | c (Confidentiality) |
| pr (Privileges Required) | ui (User Interaction) |

Reflecting the collaborative and reusable nature of ontologies, the case study is based on existing, publicly available ontologies. The Basic Formal Ontology (BFO)[33] was used as a top-level ontology. BFO provides a small core of rigorously vetted terms, philosophical principles, and strict development guidelines which may be used to develop domain ontologies. This work also used the Common Core ontologies (CCO)[34], which extend BFO to cover common things found in many domains such as information. A Cybersecurity Ontology extending the CCO lexicon was used to describe the Cyber domain. This ecosystem was then extended to describe the CVSS scoring system, as well as to introduce configuration management type notions such as version and patch numbers that are used to identify the vulnerability seeded into the case study.

Elements in the BDD (Figure 4) are stereotyped with ontological classes according to the ontologies being used in the DEFII framework. For example, the Laptop block has the <<LaptopComputer>> stereotype. These custom stereotypes will be used to map the system model to ontology-aligned data.

The cyber system is instantiated within the system model. This allows the stakeholder to observe a specific instance of a system definition and for observation of both structure (an instantiation captures structure inherited from the definition) and specific values (e.g., Internet Explorer Version 1 and Version 2). A partial instantiation table is shown in Table II.

**Table II** Partial Instance Table of Instantiated Model

| Name | ac : Attack Complexity | a : Availability | av : Attack Vector |
|---|---|---|---|
| centOS | High | Low | Physical |
| cyber System | High | Low | Physical |
| ethernet Cord | High | Low | Physical |
| Internet Explorer | Low | Low | Network |

Several tools are used to instantiate the DEFII framework for the case study. Ontotext's GraphDB triple store[35] is used as the graph repository. Dassault Systemes' SysML Authoring tool suite, including CATIA Teamwork Cloud[36] is used to create SysML models and provides remote, API-based access to the SysML elements. Protégé[37] is used for editing ontologies. Python is used for interacting with the various tools and instantiating the interfaces. The OWLREADY2[38] Python library is used for programmatic manipulation of graph data aligned to ontologies.

## 4 | RESULTS

To produce results, a DT is established (Figure 6) that uses all the interfaces to realize the defined use case. The first step is a mapping of the SysML system model from a tool specific implementation to an ontology aligned, tool-agnostic representation stored in a triple store. Second, the Direct Interface is used to directly query the graph repository, via a SPARQL query, to discover a specific vulnerability. After the vulnerability is discovered, the third step is a MATLAB analysis program pulling data specified by an MISD from the triple store via a REST API GET call. Fourth, after the analysis is complete, the triple store is updated with new values produced by the analysis. Fifth, the same REST API endpoint that was used by the MATLAB analysis program is reused by a web application dashboard to display the results of the analysis. In this dataflow, all major elements of the framework are instantiated, and the reusability of the notional Specified Model Interface is highlighted.

A functional analysis of the three success criteria identified in the introduction must be performed to determine in the DEFII framework fulfills its purpose.

**Figure 6** Digital Thread Across Different Interfaces

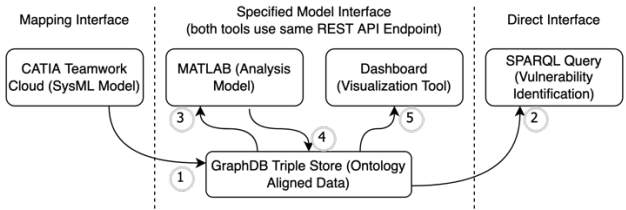

## 4.1 | Mapping Interface Instantiation

The use of a SysML System Model requires a mapping interface. Specifically, it requires a tools specific interface for the CATIA Cameo Teamwork Cloud (TWC) SysML authoring tool. The mapping makes use of a combination of SysML language elements, like the custom stereotype, and TWC specific features, like the Application Programming Interface (API) to TWC data that results in a JSON representation of the data that can be analyzed and mapped to ontology-aligned data. Mapping begins by transforming this API data to a representation in triples. SPARQL, a query language for semantic languages, can then be used to extract data relevant to the ontologies. Below is an example of a mapping rule executed in the mapping process. The example is presented in pseudocode (Figure 7) followed by a general explanation.

**Figure 7** Pseudocode describing the implementation of a mapping rule

```
"Step 1":  "Execute SPARQL Query for stereotyped
elements; store results for class and name",
"Step 2":  "Loop through the results evaluating for
'name' variable existence in defined ontologies",
"Step 3":  "Where 'name' exists as a class in the
ontology, perform the following three tasks",
"Step 3a": "Create a triple in form ([class]_entity,
rdf:type, [name])",
"Step 3b": "Create a triple in the form ([class]_spec,
rdf:type: DirectiveInformationContentEntity)",
"Step 3c": "Create a triple in the form ([subject from
Step 3b], Comm:prescribes, [subject from Step 3a])",
"Step 4":  "Put newly created triples into the graph
database",
"Step 5":  "Add a link between relevant elements in
mapped repository and original data repository"
```

First a SPARQL query is run (Figure 8). This simple SPARQL query extracts classes from the graph repository that have been stereotyped by various names. The results are then checked against known classes in the ontologies loaded. If the variable '?name' is the name of an ontological class, then the variable '?class' is mapped to the ontology aligned data. An entity and spec are created and related to each other to adhere to the way that the CCO describes information. Finally, a link back to the original, unmapped graph representation of the tool data is created to enable pushing updates back to the mapped tool. If the variable '?name' does not correspond to an ontological class, the result is discarded, and the mapping program moves to the next result. More details about this mapping process can be found in Bone et al.[7]

**Figure 8** A SPARQL Query to discover stereotyped classes

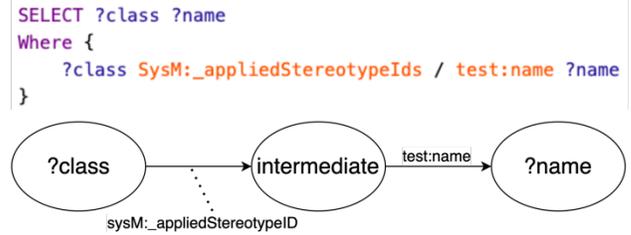

This example mapping rule is implemented as part of a collection of mapping rules that address other aspects of the SysML model representation as presented by the Teamwork Cloud tool to include other facets of the language such as instances and enumerations. Together, these rules provide a process for transferring a SysML model representation to a tool-agnostic, ontology-aligned graph data structure. While this specific result relates to the cyber case study described in the Methods section, the mapping rules discussed are more general and can be applied to any SysML model stored in a TWC instance. Thus, the mapping interface is tool-specific, but it can be used across models and domains.

## 4.2 | Direct Interface

Accessing the triple store via the Direct Interface allows for use of the SWT stack. Step 2 in the DT (Figure 6) uses a SPARQL query to directly query the triple store in search of a specific vulnerability. Seen in Figure 9, SPARQL is selective enough to isolate a specific vulnerability tied to a particular version and patch number of the Internet Explorer web browser (a contrived vulnerability created for demonstration purposes).

**Figure 9** Top: SPARQL Query; Bottom: Graph View

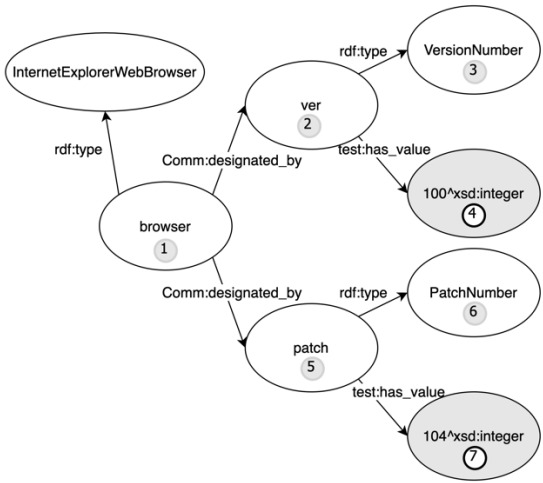

**Figure 10** CVSS Model Interface Specification Diagram (MISD)

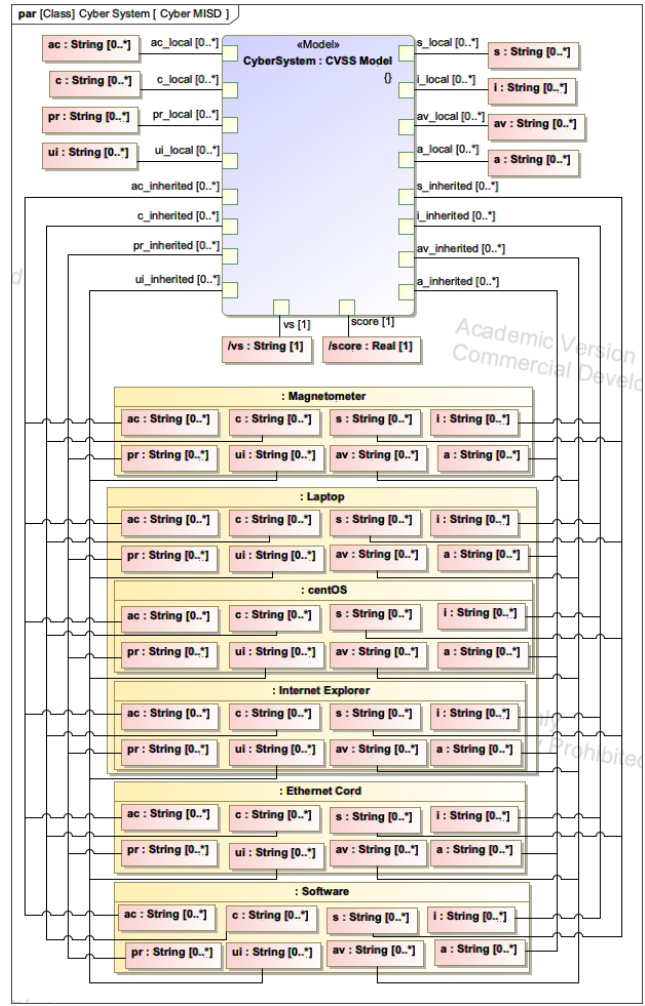

### 4.3 | Specified Model Interface with MISD

Steps 3 through 5 of the DT (Figure 6) require the use of the Specified Model Interface. In order to instantiate the interface, an MISD is created to define the CVSS model for a system level vulnerability score (Figure 10). The MISD connects parameters from a variety of levels of the architecture to a single analysis model. It also shows multiple like parameters coming into the same port. The multiple levels of hierarchy show the flexibility of the interface specification – as long as the parameter is specified within the system model, it can be attached to a specified interface for exposure. Multiple parameters sharing a single port in the analysis model collects the parameters into an array that can be analyzed as a block of data.

This interface definition is mapped to ontology-aligned data via the same mapping process described above and delineated by the <<Model>> stereotype.

Once it has be mapped to ontology-aligned data, the instantiated data associated with the definition can be transformed to a REpresentational State Transfer (REST) API. REST APIs allow stateless interaction with web services via HTTP requests. The DEFII REST API allows the Python implementation of the specified model interface to be exercised remotely via POST, GET, and PUT requests to instantiate, retrieve, and modify data in the triple store (Figure 11).

**Figure 11** Partial Results of GET Request of MISD

```
"individual": "http://testontology.org/cyber_mapped/48165ac2
a205bac33c1d_entity",
"CVSS_Model": {
    "score": 1.6,
    "s_inherited": [
        "Changed",
        "Unchanged",
        "Unchanged",
        "Unchanged",
        "Unchanged",
        "Unchanged"
    ],
    "pr_inherited": [
        "High"
```

For the CVSS model defined, a simple MATLAB analysis model was deployed to determine a system wide CVSS score along with a text-based vector for characterizing the CVSS score. Using MATLAB's *webread* and *webwrite* functions to access the REST API endpoint, data specified by the MISD is read into the analysis program, transformed by the analysis, and written back to the triple store. In this process, the data is kept in a semantically aware position – all data using the Specified Model Interface is specified and modified in terms of its place in the ontology-aligned data.

The results can be visualized using the same REST API endpoint accessed by the MATLAB analysis program. Since the interface is tool-agnostic and specified for a defined model, a simple dashboard can be created to pull the results data from the same interface. Figure 12 shows a resulting CVSS score and vector after the MATLAB analysis model has been run.

**Figure 12** Dashboard showing CVSS Analysis Results

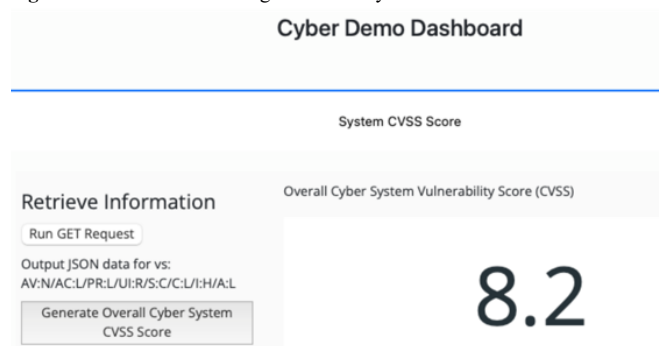

### 4.4 | Additional SWT Transformation

Additional SWT Transformation is seen in the use of automated reasoning supported by the triple store used. In the case study, Ontotext's GraphDB triple store was used. The chosen reasoning profile for this paper was RDFS-Plus, which includes sub-classes and property inferences plus transitivity. This profile was chosen because it allows for relatively fast query answering, and this application does not require more sophisticated OWL semantics. Figure 13 shows that 36,674 of the total statements in the mapped repository were inferred compared to the 19,720 statements that were explicitly provided (an expansion ratio of 2.86). This result demonstrates that additional information was inferable through automated reasoning on the ontology definitions and provided instance data.

**Figure 13** GraphDB display of explicit and inferred statements

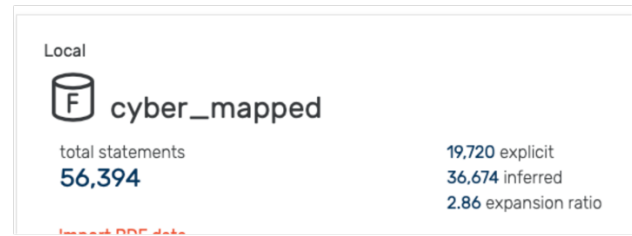

## 5 | DISCUSSION AND FUTURE WORK

### 5.1 | Analysis of Results

The introduction identifies three functional attributes of a framework to be used as success criteria for its operation in the DE context:
1. Provide clear avenues for mapping data from engineering design and analysis models to an ontology-aligned data store
2. Allow access to contained data in a flexible, tool-agnostic manner
3. Allow for the transformation and enhancement of data using various semantic technologies.

All three criteria are met by the DEFII framework presented in the paper.

Success Criteria 1 is fulfilled by the mapping interface. It provides clear guidance for what function the interface performs for the framework, and an example of its usage mapping a SysML model from an authoring tool to ontology-aligned data provides results that demonstrate the interface and give details to aid in reproduction of the interface.

Success Criteria 2 is fulfilled by the Specified Model Interface and the associated MISD. Use of the MISD allowed creation of a REST API endpoint that exposed ontology-aligned data to external tools. The results show the tool-agnostic nature of the data by accessing the same interface with two different tools, MATLAB and a web application dashboard. The MISD can also be created by system modelers that are unfamiliar with the underlying ontologies used by the DEFII framework. This enhances the functionality of the framework by expanding its usability beyond those adept in semantic technologies.

Success Criteria 3 is fulfilled by both the Direct Interface and the reasoning layer of the DEFII framework. The results show a SPARQL query accessing the ontology-aligned data to identify a vulnerability embedded in the data. The reasoning layer is demonstrated through the additional inferred statements based on the reasoning profile setup in the triple store.

The DEFII framework's use of the MISD and Specified Model Interface also promote tool interoperability. Tool interoperability denotes the ability to use multiple tools to perform similar functionality on a single model. In the cyber case study, the CVSS analysis was performed by a MATLAB program. However, a program written in Java,

Python, or a myriad of other tools and programming languages that can call REST services could perform the same analysis using the same interface. Further, another Specified Model Interface could be instantiated to provide the model specified by the MISD in a format other than the REST API endpoint. For example, it may be more beneficial to create and ingest CSV files for a particular model. As many different tools can read csv files, the potential for creating tool-interoperable data increases. The more complicated the analysis, the harder tool interoperability may become on the tool side, but the DEFII framework establishes a standard way of specifying and exposing data in a tool agnostic format that promotes tool interoperability.

**5.2 | Limitations and Future Research Opportunities**

The MISD presented in the cyber case study includes many like elements manually connected to a cyber analysis model. While the diversity of elements and parameters (multiple levels of hierarchy, single and multiple inputs to various ports) demonstrates key features of the MISD and thus is useful for presenting the overall notional interface, the actual analysis being performed could be characterized as a pattern to greatly simplify the specific instantiation of the interface. A roll-up pattern like the one presented (also consider weight[18] and cost) is recursive in nature, where like elements at one level of architecture are "rolled up" to their parent element, which then serves as an element of analysis for the next level of architecture. Future research needs to determine how to account for these types of analysis in the interface specification.

The Specified Model Interface is the preferred interface in the presented framework for most uses in a DE environment. It enables tool-agnostic implementation of a DT across multiple toolsets and consolidates interface development in a way that preserves the semantic underpinnings of the approach. It may be extended to provide tool specific data through the use of middleware that transforms the tool-agnostic representation of an interface (i.e., JSON from a REST API endpoint) to a tool specific format of an external tool that does not have web services capability. This implementation could have two benefits: developers don't need to understand SWT to create the middleware whereas they would need that background information to create a custom mapping interface, and (2) the interface could still be used by other tools. For example, an analysis program may make use of middleware to transform standard JSON to a tool specific representation, but a dashboard could access that same interface to do reporting and visualization of the data the analysis program is using/producing. If this is consistently used in practice, the Mapping Interface may be limited to the unique semantic mapping needs of a system model.

The Direct Interface enables a wide range of applications and further exploration of the SWT stack. Results in this paper show a simple SPARQL query to demonstrate access to the SWT, but more complex applications of the SWT could provide deeper functionality to the notional interface. For example, the use of the W3C recommended Shapes Constraint Language (SHACL)[39] could allow for certain verification and validation tasks to occur on the semantic representation of the system model, such as checks for well-formedness and consistency.

While this paper only integrates a single analysis model and visualization tool, most cyber physical systems would need multiple simulation-based analyses. Therefore, multiple MISDs would be linked together into a broader Assessment Flow Diagram[31] where various discipline specific simulation models (Computational Fluid Dynamics, Finite Element Analysis, Computer Aided Design, etc.) have one or more shared, interrelated parameters. Co-mingling the model of analysis with the system or mission model allows designers to relate the metadata and results from various analysis models to the system or mission level performance measures.

**5.3 | Framework Impact on Digital Engineering Domain**

Ultimately, the DEFII framework guides engineering organizations in transforming domain specific models into a knowledge representation that both provides needs for existing workflows (integrating with domain models) and establishes the foundation for additional applications depending on an integrated view of the system as a whole (Digital Assistants, reasoning, constraint checking, etc.). This forward-looking component of the DEFII framework adds value to its use in the present day as it solves an existing problem (robust data integration across multiple models) with a solution that presents opportunity beyond the current need.

Augmented Intelligence applications such as Digital Assistants can be a force multiplier in a context where solutions are becoming increasingly complex and quicker turnaround times are expected. A robust, machine readable knowledge representation of the system under design, along with the relevant domains, is needed to inform an augmented intelligence agent, and semantic technologies are a viable candidate for this representation[18,40]. Existing research into the use of Machine Learning algorithms applied to ontological data could also be leveraged to provide added value[41]. The DEFII framework gives structure for using SWT in the DE context and opens these opportunities in the future.

**6 | CONCLUSION**

The DEFII framework addresses integration and interoperability challenges in the Digital Engineering context through the use of ontology-aligned data that is exposed through to external toolsets through three types of interfaces: the Mapping Interface, the Specified Model Interface, and the Direct Interface. It introduces the Model Interface Specification Diagram as a mechanism for defining interfaces that align with ontologically relevant data without the need for the interface designer to be an expert in ontologies or semantic technologies. By taking advantage of the formal nature of ontologies and the various technologies that have been developed to enhance and use ontologies, the framework both provides for the integration and interoperability needs of model-based design and analysis today and sets a foundation for further innovation in the future.

**ACKNOWLEDGEMENTS**

This research was sponsored by the Systems Engineering Research Center (SERC), a University Affiliated Research Center (UARC) housed at Stevens Institute of Technology.